\documentclass{article}

\usepackage{microtype}
\usepackage{graphicx}
\usepackage{subfigure}
\usepackage{booktabs} % for professional tables

\usepackage{hyperref}

\usepackage[accepted]{icml2023}

\usepackage{amsmath}
\usepackage{amssymb}
\usepackage{mathtools}
\usepackage{amsthm}

\usepackage[capitalize,noabbrev]{cleveref}

\theoremstyle{plain}

\theoremstyle{definition}

\theoremstyle{remark}

\newcommand{\Collect}{Collect~}
\newcommand{\Augment}{Augment~}
\newcommand{\Compress}{Compress~}
\newcommand{\Train}{TraIn~}
\newcommand{\algoName}{\text{CACTI }}

\usepackage[textsize=tiny]{todonotes}

\icmltitlerunning{CACTI: A Framework for Scalable Multi-Task Multi-Scene Visual Imitation Learning}

\begin{document}

\twocolumn[
\icmltitle{
    CACTI: A Framework for Scalable Multi-Task Multi-Scene \\ Visual Imitation Learning   }

% It is OKAY to include author information, even for blind
% submissions: the style file will automatically remove it for you
% unless you've provided the [accepted] option to the icml2023
% package.

% List of affiliations: The first argument should be a (short)
% identifier you will use later to specify author affiliations
% Academic affiliations should list Department, University, City, Region, Country
% Industry affiliations should list Company, City, Region, Country

% You can specify symbols, otherwise they are numbered in order.
% Ideally, you should not use this facility. Affiliations will be numbered
% in order of appearance and this is the preferred way.
\icmlsetsymbol{equal}{*}

\begin{icmlauthorlist}
\icmlauthor{Zhao Mandi}{yyy,intern}
\icmlauthor{Homanga Bharadhwaj}{cmu,comp}
\icmlauthor{Vincent Moens}{comp}
\icmlauthor{Shuran Song}{yyy}
\\ 
\icmlauthor{Aravind Rajeswaran}{comp}
\icmlauthor{Vikash Kumar}{comp} 
%\icmlauthor{}{sch}
%\icmlauthor{}{sch}
\end{icmlauthorlist}

% \icmlaffiliation{yyy}{Department of Computer Science, Columbia University, New York, NY, U.S.A.}
% \icmlaffiliation{comp}{\\ Meta AI}
% \icmlaffiliation{cmu}{\\ Robotics Institue, Carnegie Mellon University, Pittsburgh, PA, U.S.A.}
% \icmlaffiliation{intern}{\\ Work done at Meta AI}
% % \icmlaffiliation{sch}{School of ZZZ, Institute of WWW, Location, Country}

% \icmlcorrespondingauthor{Zhao Mandi}{mandizhao@cs.columbia.edu}

\icmlaffiliation{yyy}{Columbia University}
\icmlaffiliation{comp}{Meta AI \\ Project webpage: \url{cacti-framework.github.io} \\}
\icmlaffiliation{cmu}{Carnegie Mellon University }
\icmlaffiliation{intern}{Work done at Meta AI \\} 
\icmlcorrespondingauthor{}{mandizhao@cs.columbia.edu}

% \icmlcorrespondingauthor{Firstname2 Lastname2}{first2.last2@www.uk}
% You may provide any keywords that you
% find helpful for describing your paper; these are used to populate
% the "keywords" metadata in the PDF but will not be shown in the document
\icmlkeywords{Imitation Learning, Robot Learning}

\vskip 0.3in ]

% this must go after the closing bracket ] following \twocolumn[ ...

% This command actually creates the footnote in the first column
% listing the affiliations and the copyright notice.
% The command takes one argument, which is text to display at the start of the footnote.
% The \icmlEqualContribution command is standard text for equal contribution.
% Remove it (just {}) if you do not need this facility.

\printAffiliationsAndNotice{}  % leave blank if no need to mention equal contribution
% \printAffiliationsAndNotice{\icmlEqualContribution} % otherwise use the standard text.

\begin{abstract} 
Large-scale training have propelled significant progress in various sub-fields of AI such as computer vision and natural language processing. However, building robot learning systems at a comparable scale remains challenging. To develop robots that can perform a wide range of skills and adapt to new scenarios, efficient methods for collecting vast and diverse amounts of data on physical robot systems are required, as well as the capability to train high-capacity policies using such datasets. In this work, we propose a framework for scaling robot learning, with specific focus on multi-task and multi-scene manipulation in kitchen environments, both in simulation and in the real world. Our proposed framework, CACTI, comprises four stages that separately handle: data collection, data augmentation, visual representation learning, and imitation policy training, to enable scalability in robot learning . We make use of state-of-the-art generative models as part of the data augmentation stage, and use pre-trained out-of-domain visual representations to improve training efficiency. Experimental results demonstrate the effectiveness of our approach. On a real robot setup, CACTI enables efficient training of a single policy that can perform 10 manipulation tasks involving kitchen objects, and is robust to varying layouts of distractors. In a simulated kitchen environment, CACTI trains a single policy to perform 18 semantic tasks across 100 layout variations for each individual task. We will release the simulation task benchmark and augmented datasets in both real and simulated environments to facilitate future research.

\end{abstract}

\section{Introduction}
\label{1-intro}
\begin{figure*}[tp!]
    \centering
    \includegraphics[width=\textwidth]{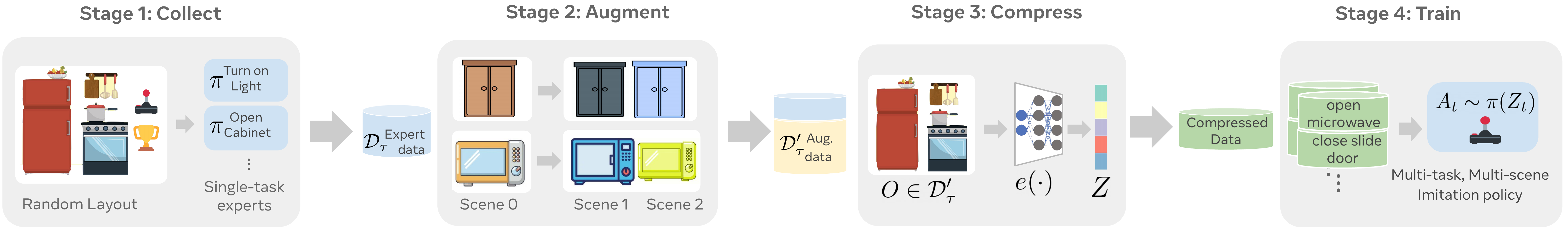} 
    \caption{\textbf{Framework overview.} Schematic of the proposed framework, \algoName's four stages. In stage 1, expert demonstrations are collected by task-specific RL agents that are trained to convergence in respective tasks. In stage 2, we augment stage 1 data by replaying these trained policy rollouts  in scenes with different visual diversity, and layout of distractor objects. This allows us to significantly multiply the data with interesting variations, beyond what is collected by limited in-domain experts. In stage 3, we learn compressed visual embeddings, and in stage 4, we train a single policy across all scene variations and tasks,  and roll it out on environments with unseen variations. 
   } 
    \label{fig:main_system}
    % \vspace{-0.7cm}
    \vspace*{-10pt}
\end{figure*} 
Despite recent advancements in learning-based control, creating an embodied agent with human-like abilities of generalizable skills remains a distant goal. Analyzing the successes of natural language processing (NLP) and computer vision (CV), it is clear that large-scale structured datasets have played a crucial role in closing the gap between human and artificial intelligence. By utilizing web-scale datasets containing high-quality images and text, the same underlying algorithms have been shown to improve results significantly~\citep{gpt2, dalle2, imagen}. However, collecting data at a similar scale in robot learning is infeasible due to operational challenges. Unlike the abundance of text and image data found on the internet, collecting demonstrations via teleoperation is both labor-intensive and time-consuming. Additionally, incorporating diversity into the data is also a challenge, as covering a wide range of objects and scenes requires vast physical resources in the case of robot manipulation.

In this work, we set out to address the above challenges by developing a framework that trains a single embodied agent to solve a wide repertoire of tasks in diverse scenes. We focus on a robot manipulation setting using visual input observations. When making design decisions with respect to data collection and policy training, a few key considerations were taken into account. First, end-to-end approaches like reinforcement learning (RL) that interleave data collection with policy learning are not ideal, as deploying suboptimal exploration policies can be costly. On the other hand, imitation learning (IL) through purely expert-collected data is also unrealistic, as covering every possible scenario for every task requires enormous resources.

In-line with the above considerations, we propose to break down the monolithic path into manageable pieces in accordance with their expense: we propose a framework, namely \algoName, that can be divided into four stages as follows: \textit{\Collect}- gather data with task specific experts, \textit{\Augment} - multiply data to boost visual diversity, \textit{\Compress} - project raw images into a low dimensional latent space, and \textit{\Train} -- learn a general multi-task agent. The \textit{\Collect}stage requires only limited collection of demonstrations, by either a human expert or a task-specific learned expert, then the \textit{\Augment}stage takes advantage of out of domain generative models to boosts the visual diversity by augmenting the data with scene and layout variations. \textit{\Compress}stage takes advantage of zero-shot visual representation models trained on out-of-domain data, which enables inexpensive training in the last \textit{\Train}stage where a single policy head is trained on frozen embeddings to imitate expert behavior across multiple tasks. Figure~\ref{fig:main_system} shows a schematic overview of the framework. We demonstrate in section~\ref{sec:method} that it is possible to instantiate this framework both in simulation and in real world using standard techniques.

Recent efforts in scaling up robot learning~\citep{jang2022bc,lynch2019play,mtopt,gato} have faced significant challenges, including the time and physical cost of running real robot hardware, and the laborious process of data collection via human teleoperation. \algoName addresses these challenges by introducing a novel data augmentation scheme that significantly reduces the need for extensive data collection, by amplifying data diversity with rich semantic and visual variations.

In our real robot setup, we utilize Stable Diffusion, a state-of-the-art image generation model~\cite{stablediffusion} that can zero-shot in-paint highly realistic object variations in the workspace, greatly reducing the need for human demonstrators to create diverse object and scene variations for each task. Unlike end-to-end approaches based on RL~\cite{mtopt} or continual learning~\cite{pnn}, \algoName ensures high data quality by utilizing experts to carefully curate small amounts of in-domain data, and the augmentation scheme enriches the size and diversity of the dataset without compromising on the quality. Additionally, by avoiding the deployment of suboptimal policies for data collection, we alleviate the need for safety heuristics that are typically required in prior work. Finally, by leveraging pretrained visual representations, we are able to scale up policy learning to large (augmented) datasets by training on compressed representations and for shorter training loops.

To summarize \textbf{our contributions}, this work presents a framework for large-scale multi-task and multi-scene visual imitation learning with the following contributions: 1) fast data collection with in-domain experts, 2) multiplication of visual diversity with realistic augmentations, 3) efficient single visual policy learning that generalizes across diverse task and scene variations, 4) a framework for leveraging large models trained on out of domain internet scale datasets to accelerate Robot Learning, 5) multi-layout multi-task simulation framework with different benchmarks that are open-sourced to the research community. Visualizations of our setup and results can be found at \url{cacti-framework.github.io}
% \url{https://sites.google.com/view/cacti-anonymous/home}

\begin{figure*}[t]
    \includegraphics[width=\textwidth]{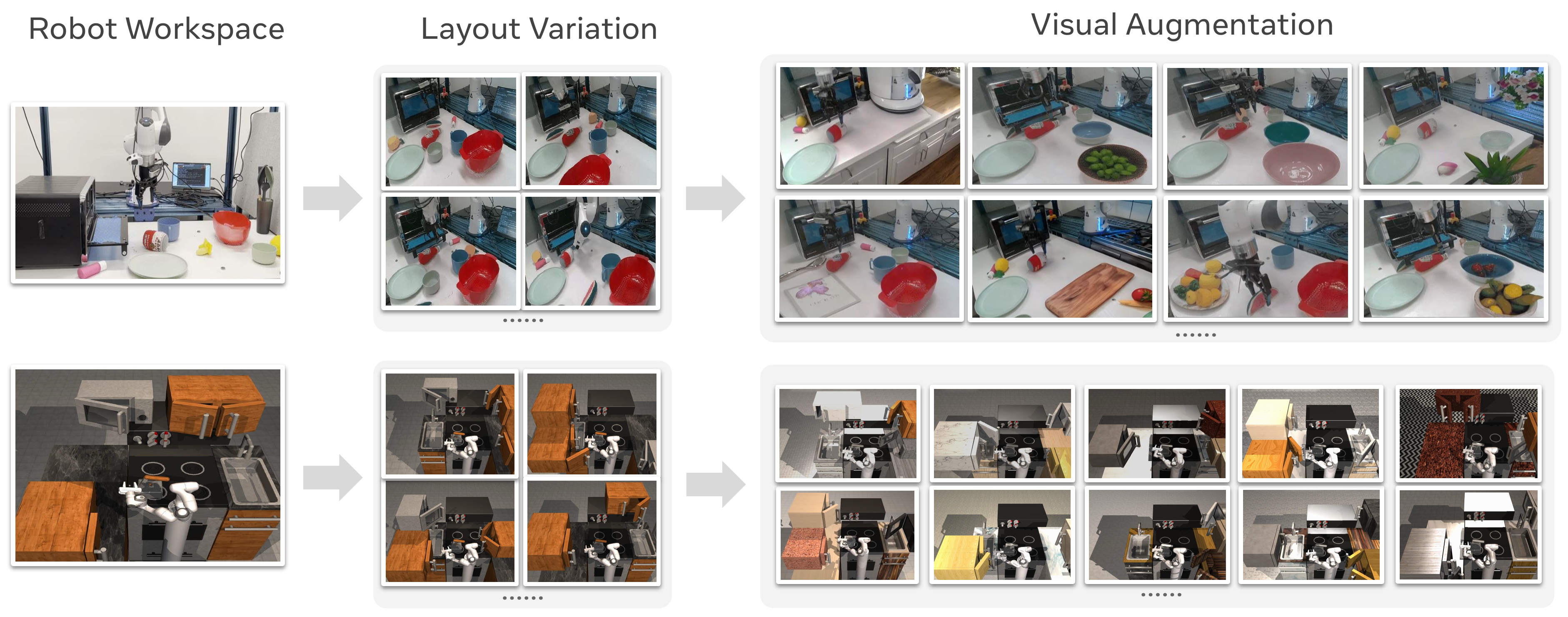} 
    \caption{We show layout variations for stage 2: Augment, instantiated in both the simulated kitchen and the real environment. For the simulation, we spawn different objects in the scene like the microwave, cabinets etc. in different locations. For the real environment, we use a novel in-painting approach to mask out locations of the scene containing objects, and in-paint the region with different plausible objects based on text prompts to Stable-Diffusion~\cite{stablediffusion} }
    \label{fig:real_sim_variations}
\end{figure*}

\begin{figure*}[t]
    \centering 
    \includegraphics[width=0.8\textwidth]{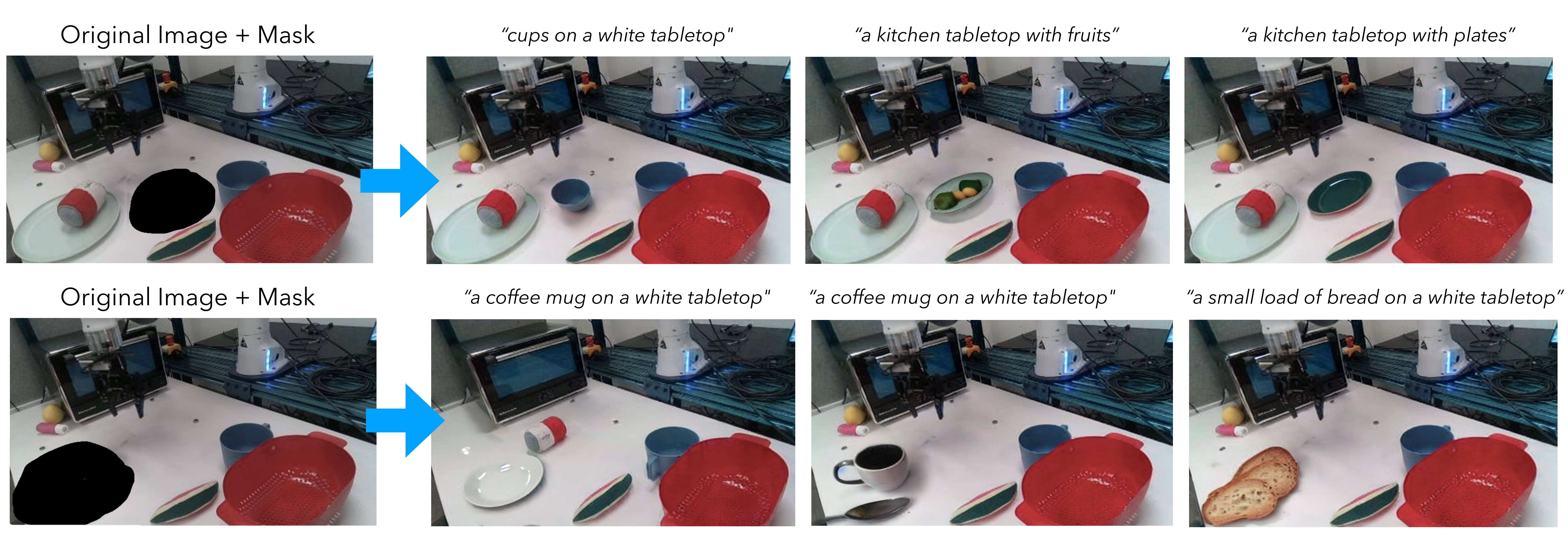} 
    \caption{\footnotesize \textbf{In-painting augmentations.} Visualization of automatic data augmentation based on controlled generation on a scene from our real-robot environment. We specify a region of the image to be edited (a mask), and a text prompt, and sample several resulting model generations. 
    We use the latest stable-diffusion model \cite{Rombach_2022_CVPR} that is fine-tuned specifically for image in-painting.}
    \label{fig:stable} 
\end{figure*} 

\section{\textbf{A Framework for Multi-Task Multi-Scene Visual Imitation Learning}}
\label{sec:method}
In this section, we provide a detailed introduction to the proposed \algoName framework, including explanations on the implementation and justifications for system design choices. 

% \VK{Before diving straight into the 4 components, it might help to prime reader why there is a need. A way to do that will be start with a desire-data for a MT-MS general agent before going into the overview section}

% \noindent{\textbf{Design considerations}} 
\noindent{\textbf{Overview.}} 
\algoName involves four stages, as illustrated in Fig.~\ref{fig:main_system}: \textit{\Collect}- gathers limited in-domain demonstration data with task specific experts, \textit{\Augment}- boosts the visual diversity across the collected trajectories using visual generative models, \textit{\Compress}- projects raw image observations to low dimensional latent embeddings using pretrained visual backbones; \textit{\Train}- trains a general multi-task policy on the augmented dataset using compressed representations as input. The four stages take into several key considerations in scaling up robot learning. 
% \VK{Order the following arguments in the same order as 4 phases. They are somewhat reversed - starts with policy and goes later into collecyion and augmentation} 
The choice to distill from single-task datasets into a multi-task imitation policy is in contrary to end-to-end learning, such as multi-task RL, which is highly resource demanding and challenging to balance across numerous tasks and learn visual representations simultaneously with control. Choice of frozen representations further allows us to work with compressed latent vectors instead of full visual frames which allows \textit{\Train} phase to recover policies using large batches and data diversity, thereby boosting generalization. Another main concern is the safety and operation cost for data collection, which is especially expensive on real robot setup with human demonstrators --- this calls for the need of Augment stage, i.e. post-factum augmentation that can leverage pretrained image generation models, and multiplicatively increase the dataset diversity at a low cost.

\noindent\textbf{Framework instantiation.} 
% TODO: 
% \VK{We introduced the framework in the last section. The next argument needs to be -- if our proposal is reasonable. What we need to argue in this section is -- Are all phases reasonable? Can all phases can be instantiated in common research paradigms -- both in sim and real? Even in complex setting such as general kitchen (note we had a reviewer criticism on why kitchen only), let alone other simpler cases. What follows here very descriptive of what we did. We need to motivate it from the POV of why it all make sense. And what's our strategy to present this proof.}
Because of their ubiquity in real-life manipulation scenarios, we choose to present and validate \algoName in kitchen scenes, which involve everyday household objects such as cabinets, drawers, toaster ovens, fruits, mugs, strainers and plates. We set up both a real robot workspace and a simulation kitchen environment, as visualized in Fig.~\ref{fig:real_sim_variations}.
% There are third-person RGB cameras in the workspace that provide different views of the same scene. 

\textbf{In the real robot setup}, we set up a toy kitchen tabletop with a Franka Emika Panda arm and the toy objects as shown in Fig.~\ref{fig:real_sim_variations}. The Franka arm has a parallel gripper, and is operated with joint position control with 8 action dimensions and 12.5 Hz control frequency. We define 10 tasks that involve manipulating the tabletop objects, each task is constrained to one single stage (single goal) and provides additional conditioning for learning multi-task policies. See Appendix \ref{fig:10task} for full description of the 10 tasks and other setup details. 
% \VK{What's the equivalence of sim layout variation in real isn't articulated well. We need some way to establishing that there is some equivalence between the layout variations and real world distractor objects. I like how the figure 2 is organized as layout variation + visual augmentation}

\textbf{In simulation}, we create a simulated kitchen that supports 18 semantic tasks (e.g. open microwave, close slide-cabinet, etc) and randomly-generated layout variations 
% \VK{Show figure}. 
Each layout has a different arrangement of the main kitchen objects (for example, placing the microwave next to the sink v.s. in the top shelf next to the cabinet). 
% \VK{9/11: technical arguments behind various details we are presenting below is missing. Explicate what's the technical merit of having noisy actions, augmentation, semantic augmentation, out of doamin rep, fine tuning, etc.}

\subsection{\textbf{Collect: Small in-domain expert data collection}}
% \VK{9/11: Add high level (sim/real agnostic points) as intro. Collect is more expensive. The goal for the collect phase is to seed the process with minimal expense. etc etc.} 
The goal of this stage is to collect a limited amount of expert demonstrations, while minimizing the cost of data collection both in terms of human labor (tele-operationg real robot) and computational cost (training RL experts in simulation).  

\subsubsection{\textbf{Real-robot environment}} 
  We choose kinesthetic teaching by a human expert to collect demonstration trajectories --- this has the benefit of recording once and replaying the same action sequences several times more, which allows further variations of the scene layout. and expert demonstrations are collected in a single-object, single-task setting. After the human expert hand-holds the robot arm and to complete a task, the joint poses and end-effector information of the robot at each time-step of the trajectory are saved. For each of the 10 tasks, the demonstrator collects 5 trajectories of kinesthetic demonstrations, each is replayed 20 times, and at each time of replay, the demonstrator re-arranges the non-target objects. 

\subsubsection{\textbf{Simulated Kitchen}} 
% \VK{9/11: move most of these details to instantiation section above. reference to figure to paint a mental picture} Our simulated platform consists of simulated kitchen environemnts \VK{9/11:Why?} consists of several common objects like a microwave oven, light switches, a kettle on the stove, and several different types of drawers. The positions of every object in the scene can be randomized, respecting physical constraints\VK{9/11:unclear why this is important and how its being used in this phase. Perhaps we should present this in next phase}. The agent is a robot arm \VK{9/11: how may dof?} in the scene that can be moved with joint control commands.

We use a standard on-policy RL algorithm, namely NPG~\cite{npg}, to train a fleet of single-task, single-layout expert policies $\pi(s_t)$ from state-based input observations $s_t$. For each task $\mathcal{T}$, we define a reward function $r_{\mathcal{T}}$, and the expert policy $\pi_{\mathcal{T}}$ receives the current simulator state $s_t=\{robot, object\}_{pos, vel}$ as observation at time-step $t$. We generate 100 layout variations for each of the 18 tasks and trains an expert policy for each layout, hence resulting in a total of 900 policies. The expert training runs are inexpensive and easily parallelizable --- in practice, we initialize a large batch of parallel training runs and use a threshold of 90\% success rate to filter converged policies as experts.
% \VK{The layout variations in sim, and object variations in real seems disjoint choices. Is there any way we can draw an equivalence between the two. Second thing that seems a little odd is the difference in the data collection strategy between sim and real.}
% The dataset $\mathcal{D}_\tau \forall \tau$ is collected using these experts under a set of single-task settings (not multi-task). Example tasks include opening a microwave, opening the top cabinet, turning the switch etc (more details in the Appendix \ref{}). The  environment is  are 18 semantic tasks, and up to 100 layout variations, so a total of 900 expert policies are trained in this stage. 

\subsection{\textbf{Augment: Semantic scene variations for augmentation}} 
% \VK{9/11: outline  goal/ consideration/ expense of this stage? Introduce both visual and semantic augmentation first and then go into how they are instantiated for sim and real in individual sections below }
%\Homanga{9/11: note we already introduce visual and semantic augmentations in section III overview}
This stage aims to increase the diversity of the raw dataset $\mathcal{D}_\tau$ and yield the augmented dataset $\mathcal{D}'_\tau\forall\tau$, which will be used for visual policy learning.  
To do so, we introduce two types of augmentations, \textit{visual} and \textit{semantic}. Visual augmentations involve changing visual attributes of the scene, such as object texture lighting conditions. Semantic augmentations involve changing the layout of objects in the scene, namely their positions and orientations, or even adding new, artificial objects. 
% \VK{We introduce this categorization but its not as cleanly followed in the real and sim details. It will be nice to be very explicit about this categorization in the following two subsections}. Visual augmentations involve changing visual attributes of the scene, such as object texture lighting conditions. Semantic augmentations involve changing the layout of objects in the scene, namely their positions and orientations, or even adding new, artificial ones. 
% \VK{Here is where the reviewer wanted to see comparison with (a) other generative models for schematic augmentation (b) and comparison against common pixel based augmentation}

\subsubsection{\textbf{Real-robot environment}}  
% The kinesthetic replay scheme on the real robot allows semantic augmentation via manually varying different attributes of the scene, such as changing the background lighting. In addition, we incorporate three other augmentations: 1) injecting action noise during replays to ensure wider coverage of action distribution and mitigates covariate shift issues; 2) manually shuffle the positions of distractor objects across the scene, and swap some objects in and out of the scene; \VK{Given our recent pivot on the pitch -- out-of-domain  models for robot learning, this section needs an overhaul and refocus. We need to stress the importance and awesomeness of this part} 3) we propose a novel scheme for incorporating automatic semantic scene variations, \textit{without} physically modifying the objects in the scene. We use latest advances in generative modeling~\cite{dalle2,stablediffusion} that lets us perform controlled scene in-painting, which operates offline on the dataset level and doesn't require additional robot operation hours. \VK{We never make any arguments on why common pixel based augmentation isn't enough for augmentation.}

We consider the open-sourced Stable Diffusion model~\cite{stablediffusion}, and run zero-shot in-painting inference on the lab-collected robot data. The model takes as input an image of the scene, and a region for modification, specified as a binary mask. Such controlled generation lets us keep the rest of the scene unchanged, and introduce new plausible objects in the region specified. By automating this process with different text-prompts, we can obtain several visually augmented demos with zero extra human effort for data collection. Fig.~\ref{fig:stable} shows a visualization of what controlled generation looks like for a scene from our real robot environment. The generated images place highly realistic-looking objects on designated locations on the white tabletop. These augmentations provide more semantic variations in the scene, that is not possible to obtain based on traditional augmentations like color jitters, random crops etc. that are common in the robot learning community~\cite{drq,dreamer}. 

%Please refer to Appendix section~\ref{sec:stage2details} for details. TODO: stress why standard aug is not sufficient

\subsubsection{\textbf{Simulated Kitchen}} 
% \VK{It will be nice to be very explicit about the categorization of the augmentations -- visual, semantics , and use it as the unlying theme unifying the details between sim and real} 
The previously trained expert RL policies operates without visual inputs, which gives us freedom for generating visual observations. We replay an expert trajectory and vary visual attributes (i.e. color, texture, and lighting) of the scene during rendering, which generates visual observations without changing the expert policy's physical effects in the scene. We save the resulting trajectories which now contains diverse image observations that can be used to train a visual policy. For semantic augmentation, we randomize the pose of non-task (i.e. ``distractor") objects during action replay, and add action noise at each time-step, similar to the real robot setup. We render 50 episodes (50 time-steps each) for each task and layout combination, which results in a training dataset of 45,000 episodes. 
%\VK{9/15: Missing: Over all data size. Can be moved to appendix as well but mention and cite}
 
\subsection{\textbf{Compress: Representations pre-trained on internet or in-lab data}}
% \Homanga{we need to change the phrasing of this section based on frozen ood R3M not doing as well as in-domain MoCo} \VK{9/15: Real world evals are still R3M correct? Lets pivot to sim data is reasonable size, so we trained in-domain representation. Data distribution is small in real, so we leverage out-of-domain  data and then fine tune it using in-domain data. So both use in-domain data to the extent possible.}
The \Compress stage of our framework involves encoding image observations into low-dimensional embeddings, which makes it easier for the downstream policy to learn across complex semantic variations in the scene, and potentially generalize to new scenes with different attributes. This also helps to decouple representation and policy learning, and independently optimizing for each component through separate methods and architectures. We explore the use of representation networks trained with large-scale out-of-domain internet data, as well as representation models trained with only in-domain data from the simulator. For the former case, we use the R3M model~\cite{r3m} which has demonstrated strong empirical performance in various imitation learning tasks. For the latter, we train a ResNet-50 model using MoCo~\cite{moco} on the in-domain data. Parisi et al.~\cite{pvr} studied various components of self-supervised representation learning for policies and found MoCo to perform particularly well. Considering the large volume of our simulated dataset, we opt to train an in-domain representation model using MoCo to serve as performance baseline. Since data distribution in the real setup to too small to train a model from scratch and consider the R3M model trained with out-of-domain data, and fine-tune it using the in-domain data that we collect in stage 2. 
% \VK{We are mixing details of our methods here with baselines and ablations (1) We should lead with our pitch of leveraging our of domain models to accelerate robot learning (2) Lean heavily on the fact that -- in domain data is small and any representations trained in doamin will be suboptimal (b) lets discuss though if our results suggests this ? or if we can make find arguments in this favor}

\subsection{\textbf{TraIn: Multi-Task Multi-Scene Visual Policy Learning}}

% todo: give a name for the kitchen benchmark 
 
\begin{figure}[t]
    \centering 
     \includegraphics[width=0.99\columnwidth]{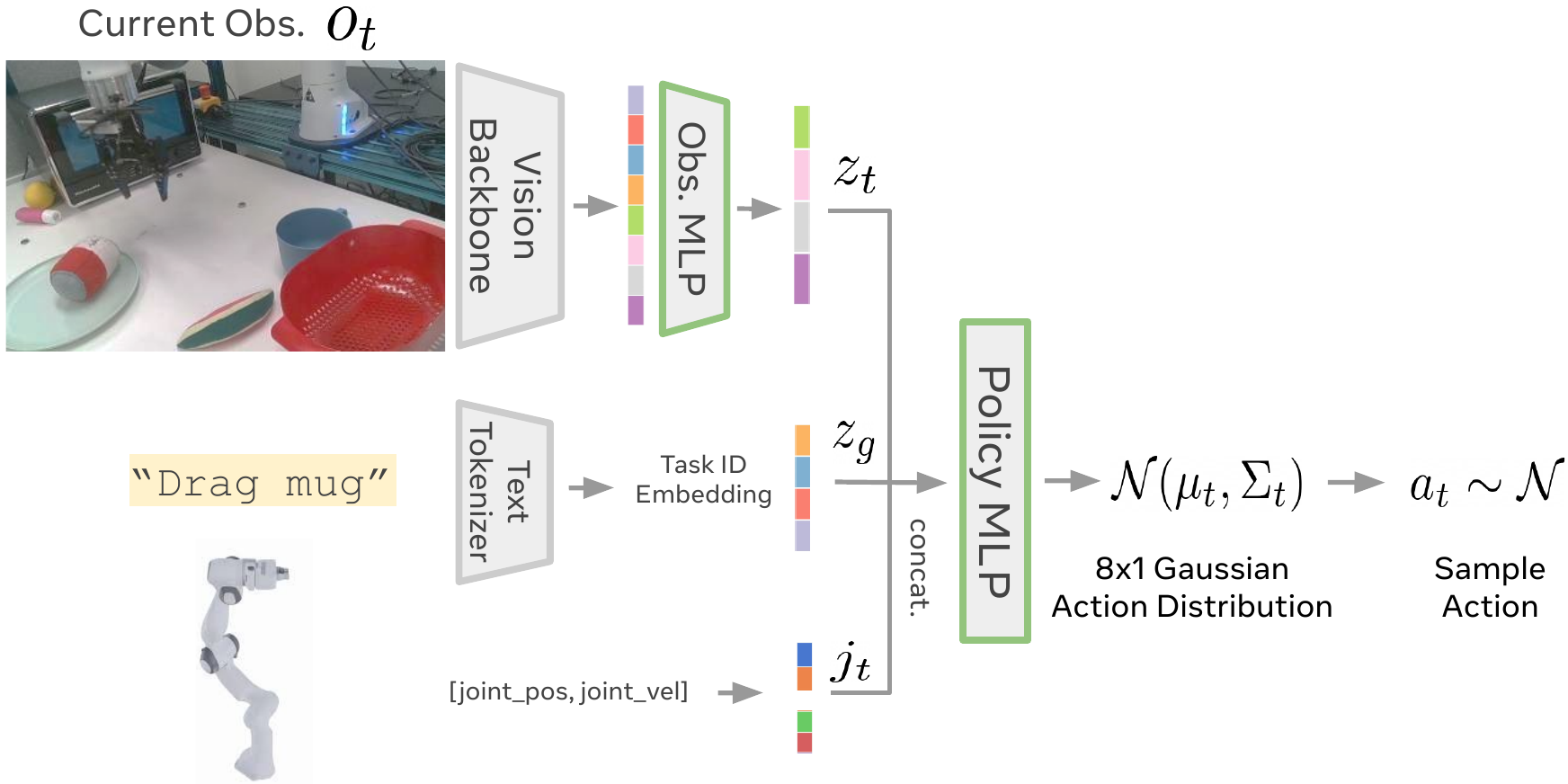}
    \caption{\textbf{Policy deployment pipeline.} Schematic of the deployment setup for the final multi-task multi-scene visual imitation policy. A task embedding is provided by the experimenter, and the predicted actions $a_t$ are executed in the environment to obtain the next observation $o_t$.}
    \label{fig:architecture}   
    \vspace*{-0.5cm}
\end{figure}

The final stage trains a single policy with the visual backbone from stage three on the entire multi-task multi-scene data, in simulation and the real environment respectively. The overall task-conditioned policy architecture and the deployment protocol during evaluation is shown in Fig.~\ref{fig:architecture}. For real robot experiments, we opt to using tokenized text embedding from each task's name string for the policy's task conditioning. The tokenizer we use is same with that used in R3M \cite{r3m} and kept the same during deployment for each task. At time-step $t$, the input observation $o_t$ and task name string are respectively embedded to latent representations $z_t,z_g$. The embeddings are concatenated and fed to an MLP, which we denote as $\pi$ and parametrized by learnable parameters $\theta$ and $\sigma$, as shown in equation below. The policy MLP outputs the mean of a Gaussian action distribution and adds a learnable $\sigma$ noise. The policy is trained to minimize the mean-squared behavior cloning loss. 
%We experiment with different variations in training, e.g. by freezing the R3M embeddings, and fine-tuning them. 
% 
$$
\hat{a_t} = \pi_{\theta}([z_t,z_g])+ \exp{(\sigma)} * z; \  z \sim \mathcal{N}(\mathbf{0}, \mathbf{1}) $$ \vspace{-0.5cm} $$ \mathcal{L}_{\text{BC}} = || \hat{a_t} - a_t ||^2 $$ 
The decomposition of \algoName into four stages, with visual IL as the last stage provides several advantages over monolithic frameworks such as online multi-task RL. From a frameworks perspective, it is difficult to learn goal-conditioned behaviors with monolithic approaches, as they require learning some auxiliary goal-sampling distribution which often doesn't capture the space of plausible interactions~\cite{pong2019skewfit,nair2018imagined}. In contrast, the stage four of \algoName can easily learn a goal-conditioned policy by considering the last observation of every augmented trajectory to be a goal image.

% \VK{pretty much end of every stage needs a paragraph like this (with appropriate subheading) outlining the clear benefits of our choices and its relative strengths amongst alternatives.}

\section{Experiments}
We set up experiments on both simulated and real-robot environments in the aim to understand the following research questions:
\vspace{-0.4cm}
\begin{itemize}
    \item How effective is \algoName in learning behaviors for diverse tasks and scenes? This effectiveness is primarily validated by how well a final policy can perform on the training tasks and variations. Additionally, we also compare the cost efficiency of \algoName with other alternatives, such as monolithic approaches.
    \vspace{-0.2cm}
    \item How do the resulting policies from \algoName generalize to novel scenes? \algoName is designed to facilitate further scaling up of robot learning, which has the promise of generalization to unseen scenarios. Hence, after ensuring a resulting policy can execute well on the training tasks, it would further reinforce the merit in our design if the policy also shows promising signs of generalization to unseen objects or layout variations.   
    \vspace{-0.2cm}
    % \VK{Our scaling laws are quite nice in sim, consider adding a claim there}
    \item How do variations in instantiation details of the different stages in \algoName impact the framework's effectiveness? Using behavior of the final policy as the main indicator, we set up additional experiments that ablate on more specific sub-parts of \algoName, in the hope to provide insights on the framework design.
        
\end{itemize}

\subsection{Real Robot Experiment Setup}
\label{exp:real_env}
\textbf{Task Setup}
A Franka Emika Panda robot arm is placed in front of a white tabletop with toy household objects, such as fruits and kitchen utensils, using which we define 10 manipulation tasks, such as dragging a mug, picking and placing various objects, and opening toaster oven. 

\textbf{Policy Training and Deployment}
Consistent with policy training, during evaluation, a task name string is provided through user input to indicate which task to perform (e.g. ``drag\_mug"), and this input text string gets embedded with a frozen tokenizer. To evaluate generalization, we randomly shuffle the locations of distractor objects (some are unseen from training) and induce organic variations in the initial positions of the target objects. Please refer to Appendix Section \ref{app:real_exp} for more details and visualizations of all 10 tasks.

\subsection{Simulation Experiment Setup}
\label{exp:sim_env} 
\textbf{Task Setup}
The environment contains 18 tasks involving 8 main objects, such as turning on a light switch, opening the left door on cabinet, turning a knob. In parallel to the 18 semantic tasks, we generate 100 random layouts that re-organizes the kitchen setup. 

\textbf{Policy Training and Evaluation}
A 43-dimensional context embedding is used to condition a multi-task policy model. During evaluation, a task considered success if the final pose of the object is within a given error threshold of the goal pose of the object. To evaluate generalization, we additionally generate 10 more layouts in the simulation for each task, and randomly re-sample visual properties. Please refer to Appendix section \ref{app:sim_exp} for more experiment details. 
   
We will open-source our large-scale augmented dataset, with 100 different semantic variations per 18 base tasks, with 50 trajectories per variation, along with simulator code that allows generating countably infinite visual augmentations. We denote this benchmark \algoName-Sim-100.
 
% Please refer to the supplementary video for a demonstration of the range of variations in our simulated kitchen setup. 
\vspace{-0.2cm}
\subsection{Ablation Experiments}
\vspace{-0.1cm}
\subsubsection{Visual Representations (Compress)} 
In \algoName, using frozen visual backbones that were trained on out-of-domain data provides significant improvement in storage and computational efficiency for policy training. However, this raises concern on possibly lossy compression in the Compress stage that may cause the policy to not receive valuable information present in the raw visual input. 

To better understand the trade-offs between training with raw visual input v.s. compressed visual embeddings, we train a version of \algoName that simultaneously fine-tunes the visual backbone during policy training. To verify whether the visual representations trained on out-of-domain data is of sufficient quality, we pre-train another frozen visual backbone on strictly in-domain real robot image data, and use the frozen embeddings as input to \algoName's final training stage.
\vspace{-0.3cm}
\subsubsection{In-painting Augmentation (Augment)}
We ablate on the role of our proposed in-painting augmentation technique in enabling robust and sample-efficient imitation learning. An ablation experiment was set up by using only standard image augmentation techniques (i.e. random crop and color jitter) on the original, un-augmented dataset for training, and we compare the performance of the resulting multi-task policy with the policy that was trained with augmented data. Note that, data used for this ablation policy has different variations on the image level, but contains the exact same amount of information in terms of expert action trajectories. 
\vspace{-0.3cm}
 \subsubsection{Comparision with End-to-End Approaches}
In simulation, we additionally run a large-scale, end-to-end reinforcement learning that takes in pixel inputs and learns across all the available tasks and layouts. The training run was not able to converge and the RL agent fails to achieve beyond zero success rate.

% \VK{A somewhat important pattern that I have observed in the writting is -- its mostly descriptive of what was done. I good write up has following flow (1) What's the objective (2) critical analysis of the objective (3) what approaches exist (4) What's our intuition/idea (4) Comparative strength of our idea (5) AND then we describe the idea in details. This write up is diving straight to (5)}

\begin{table*}\centering

\begin{tabular}{lcccccccc}
\toprule
  % &  \text{Sim10 \\ Train} & Sim10 Heldout & Sim50 Train & Sim50 Heldout \\
   & \multicolumn{2}{c}{Sim $10$}  && \multicolumn{2}{c}{Sim $50$} && \multicolumn{2}{c}{Sim $100$} \\
\cmidrule{2-3} \cmidrule{5-6} \cmidrule{8-9}
   & Train & Heldout && Train & Heldout && Train & Heldout \\ 
\midrule 
State-based & 81.2 +/- 2.2 & 14.1 +/- 3.1 && 83.3 +/- 2.2  & 31.6 +/- 4.4    &&  91.3 +/- 1.1 & 47.2 +/- 4.5   \\ 
Out-domain  & 51.3 +/- 4.3 & 9.5 +/- 2.0  && 64.7 +/- 3.6  & 30.4 +/- 5.6    &&  62.0 +/- 3.8 & 33.1 +/- 4.3   \\
In-domain   & 88.7 +/- 1.2 & 18.8 +/- 2.9 && 72.1 +/- 2.8  & 30.2 +/- 5.2    &&  75.9 +/- 2.6 & 38.4 +/- 4.7  \\
% End-to-End RL & 0           & N/A          & 0             & N/A  \\ 
\bottomrule
\end{tabular}
\caption{\textbf{Simulation results}. We compare different layout randomizations for training (10, 50, 100) and report success rates for evaluations on both layouts seen during training, and heldout layouts unseen during training. Note the increased generalization performance to unseen layouts (heldout) as we scale up the number of layout variations used in the training process. This suggests the benefits of having strong semantic data augmentations in the Augment phase. 
}
\label{sim-result}
\end{table*}

% \VK{Im assuming that we are still working on the experiments sections. Regardless, I'll outline a few points which we will hopefully get to (a) we have 3 experimental questions at the begining, we need to ensure that we have answered thm separately (b) What exactly are our baselines (c) what exactly are our ablations (d) Critical analysis of the reported results and deductive analysis } 
% \VK{Ablation: Pixel augmentation vs scemantic augmentation}

\begin{figure}[t]
    \centering      
        \includegraphics[width=0.85\columnwidth]{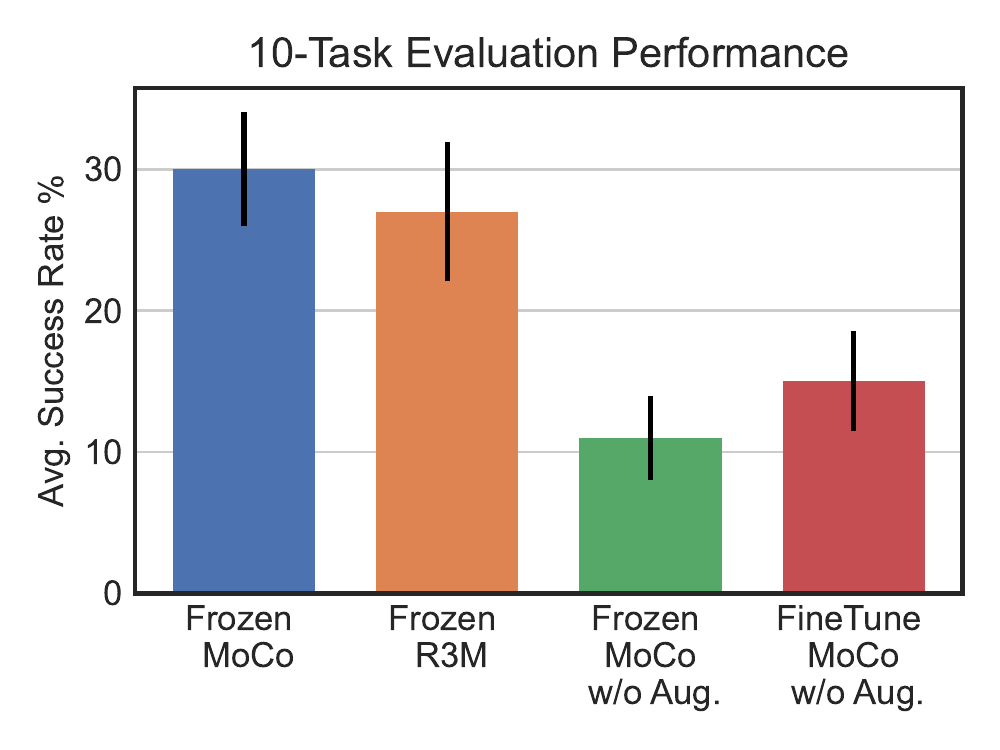}
    \vspace{-5mm}
    \caption{\textbf{Real robot evaluation results} We evaluate each compared multi-task policies and report average success rate across 10 training tasks. We remark that training with versus without the in-painting augmentation results in significant performance difference: whereas the policy can achieve decent overall success rate from frozen visual backbones trained entirely on out-of-domain data, even fine-tuning on the in-domain, no-augmented data cannot reach comparable performance. }
    \label{fig:real}    
    \vspace*{-0.5cm}
\end{figure}

\begin{figure*}[t]
    \centering   \includegraphics[width=\textwidth]{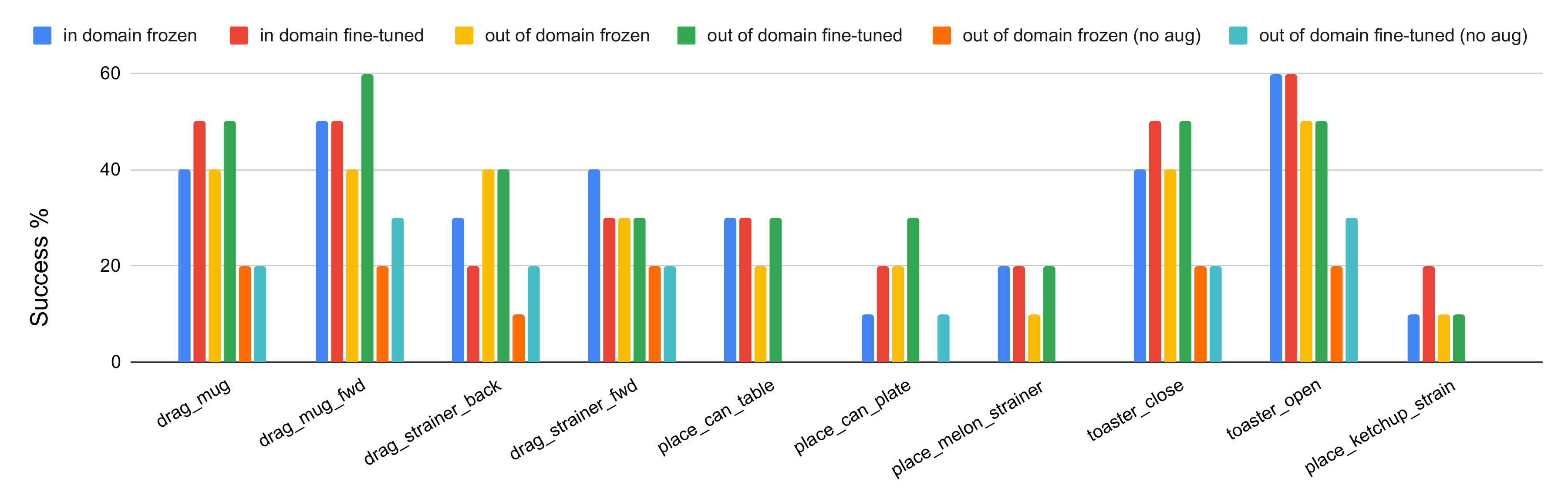}
    \vspace*{-0.6cm}
    \caption{\textbf{Real world evaluation with MoCo representations.} We report detailed results from the real robot environment tasks (10 tasks using single policies) using the evaluation setup described in section~\ref{exp:sim_env}. The different evaluated variants include training on in-domain vs out-of-domain  data, frozen representations vs. fine-tuning, and using vs. not using augmented data. 
    % \VK{Also add aggregate results over all tasks}
    }
    \label{fig:real_results}
    % \vspace*{-0.7cm}
\end{figure*}

\subsection{Results}

\textbf{\algoName successfully and efficiently trains multi-task imitation policy}
We report evaluation performances from both real robot and simulation experiments. As shown in Fig \ref{fig:real_results} and Table \ref{sim-result}, \algoName is able to efficiently train a 10-task visual policy that achieves overall $\approx$ 30 \% success rate  when deployed on the real robot, and a visual policy in simulation that achieves $\approx$ 62\% success rate across all 18 tasks and 100 variations. In contrast, end2end RL \cite{redq} achieves $0\%$ success. Although our task performances are far from perfect, the results are a promising sign that \algoName can leverage entirely frozen visual representations and be efficiently scaled to even more tasks and variations.

\textbf{\algoName enables training policies that show promising generalization.}
In real robot experiments (Fig.~\ref{fig:real_results}, our deployment involves drastically shuffling the background objects, hence the deployment success rate provides a good indicator for how well the policy can be robust to out-of-distribution distractor layouts, and also completely novel distractor objects that are introduced in the scene. In the simulation results summarized in Table~\ref{sim-result}, we see that generalization to held-out layout variations increase from Sim-10 to Sim-100, indicating that doing data augmentations for increasingly large layout variations during training significantly aids in boosting out-of-domain generalization performance. 

\textbf{(Augment stage ablation) The semantic augmentations in \algoName allow for higher success rates over traditional pixel-based augmentations} In the real-world results summarized in Fig.~\ref{fig:real} and described in detail in Fig.~\ref{fig:real_results}, we see significant performance difference of around 15-20\% absolute success rate among the variants that use semantic in-painting augmentations and those that use only standard color-jitters and random crop image augmentations (these are the no Aug variants in the plots). This confirms our hypothesis about semantic augmentations being helpful for task performance with scene variations, and also in generalization to unseen objects in the scene, that are introduced during deployment

\textbf{(Compress stage ablation) The Frozen Out-of-domain representations of \algoName are competitive with in-domain and fine-tuned representations} From the real-world results in Fig.~\ref{fig:real_results}, we see that the out-of-domain frozen representations perform almost as well as the in-domain frozen, and fine-tuned representations. This is very promising because it indicates potential for using the vast amount of internet data of images and human doing everyday things to pre-train representations for robot learning. For the simulation results in Table~\ref{sim-result}, the difference between out-of-domain and in-domain representations are a bit higher, probably because the simulation environment is \textit{visually} very different from the real-world data used for pre-training the out-of-domain embeddings. 
% \textbf{Ablation experiment results}

\section{Related Work}
\subsection{Frameworks for scaling robot learning}
 Prior works on scaling robot learning have largely focused on the RL paradigm, either through multi-task RL~\cite{mtopt} or meta-RL~\cite{reptile,pearl} and shown that shared learning among tasks amortizes the cost of acquiring diverse behaviors compared to training single policies for individual tasks~\cite{agpinto,impala,riedmiller2018learning}. The main reason for success in these settings has been that most tasks share some common structure (for example reaching and grasping behavior components), and such structures can be discovered through the learning of shared policy. %In addition, while learning from visual observations, the cost of learning representations of recurring objects and scenes can be amortized. 
 This is useful from the perspective of designing frameworks that are scalable with efficient re-use of data across tasks. 
%  \VK{9/10:A common theme in the write is -- there is a solid point (like the line before) but there is no conclusive remark presented on it. Its left for the readers to infer the consequences - positive or negative. Drive a point home, don't assume the readers to be smart to do so}.
%  Closely related to multi-task learning is the paradigm of meta RL, where an agent is trained to acquire and "average" representation across tasks, which is optimized during test-time through task-specific interactions~\cite{reptile,pearl}. 
 Recent work~\cite{mandi} has found that learning pre-trained representations and simple multi-task learning outperforms most meta RL approaches. There have been similar findings on IL from large offline datasets~\cite{gato}. \algoName is inspired by these findings where we collect offline data, and use pre-trained visual representations for multi-task IL on the offline data, but instead of collecting all the data by experts~\cite{gato} (which is expensive in robotics), we have an efficient data augmentation scheme for multiplying a small set of expert data. In the next paragraphs, we discuss \algoName's four stages in relation with respective prior works.
\subsection{Visual policy learning.}  
% Talk of prior work in toyish environments like dm control, or in non realisitc simulators like meta-world. we can cite several of ours and co-authors++ papers as well for bashing (because icra is single blind) Several of these method rely on online learning in the particular environment (RL methods) and require very clean task-specific data without scene complexity. 
% We can cite papers that do domain randomization in sim, and point out that they are limited by the extent of randomization. 
%\VK{Flow leading to this point needs improvement. At this point reader isn't clear why we are talking about visual policy learning. How does this connect with the landscape of the proposal} 
Learning control policies from visual observations helps amortize the cost of learning representations of recurring objects and scenes~\cite{visual1,visual2,visual3,dreamer,visual_imitation1,visual_imitation2,visual_imitation3}. However several prior works have looked at visual policy learning in simple simulated environments like the DM Control Suite~\cite{dmc} that involves stick agents locomotion~\cite{dreamer} or in simplified manipulation environments such as Meta-World that involves only a few objects in the scene with a robot arm~\cite{metaworld,visual_metaworld}. Other works have tackled policy learning in much more complex settings like a simulated realistic looking kitchen with several objects, but assume ground-truth simulator state observations instead of visual inputs~\cite{relay,bet}. In contrast, \algoName(sim) is based on a simulated kitchen similar to~\cite{relay} but with much more diversity of visual observations and layouts, and incorporates only visual observations as inputs to the multi-task multi-scene agents making it readily amenable for real-world environments where it is not possible to obtain ground-truth states of objects in the scene. %This also makes our framework readily amenable for real-world environments with several objects where it is not possible to obtain ground-truth states of objects in the scene.
\subsection{Domain randomization.}
 Domain randomization~\cite{DomainRand,Sim2RealJamesDJ17, dynamicsRand,DomainRand1, DomainRandLerrel, Tzeng2015TowardsAD, Sim2RealSadeghiL16} bridges the reality gap by leveraging rich variations of the simulation environment during training. The hope is that by adding random variability in the simulator, the real data distribution will be within that of the training data.  This has been useful in recent advances for visual navigation and manipulation in real-world environments \cite{James_2019_CVPR}. Inspired by similar ideas, we go beyond simple domain randomization like color jitters, camera motions, texture changes, to more semantic augmentations based on distractor objects, and layout variations, through hindsight relabeling of limited expert demonstrations. 
 \subsection{Pretrained models for robot learning}
 We incorporate a novel image in-painting~\cite{inpainting} based data augmentation that lets us add different realistic objects in the scene by running inference through trained generative models~\cite{dalle2,stablediffusion}. It is only until recently that these diffusion-based image generation models achieve state-of-the art results, and to our best knowledge, we are the first to utilize such pretrained models for zero-shot generation on robotic data --- this is highly encouraging for the robotics community, as we have verified that these models can be useful even on data from strictly unseen lab settings. Concurrently, \cite{kapelyukh2022dall} proposed to use DALL-E model \cite{dalle2} to generate image goals for object rearrangement.
\vspace*{-0.2cm}
\subsection{Representation learning for control.}   
Recent progress in video prediction and self-supervised learning, such as developing suitable lower bounds to mutual information (MI) based objectives~\cite{e2c, svvp,visualforesight, dreamer,ha2018world, planet, lsp,slac,gregor2019shaping}, have enabled learning of visual representations that are useful for downstream tasks. Prior work have examined pretraining on large datasets like ImageNet~\cite{imagenet} and Ego4D~\cite{ego4d}, and using the frozen representation for doing downstream robot control ~\cite{pvr,r3m}. \algoName leverages such frameworks for learning compressed visual representations, both with out-of-domain internet data of human videos, and with in-domain augmented dataset that is generated as part of the framework. 
\vspace*{-0.2cm}
\section{Discussion and Conclusion} 
\vspace*{-0.2cm}
We developed a framework for multi-task multi-scene visual imitation learning, and instantiated it both in simulation and in the real world. Our framework incorporates several components like fast and efficient data collection, novel data augmentation, compressed visual representations, and a single control policy trained over augmented datasets. We demonstrate efficacy of the framework in a large-scale simulated kitchen environment with several variations in the tasks, type of objects, and randomizations in the scene. We build the same framework in a real-world robot environment that includes several everyday objects found in common kitchens. We show the efficacy of novel augmentations like in-painting images in a dataset with different objects based on prompting from out of domain deep generative model. Our results strongly indicate that large (generative as well as representational) models trained on diverse internet scale out of domain data sets can be leveraged to boost generalization in robot learning where in-domain data collection face fundamental challenges. Building on our framework, we believe future work would involve developing architectures capable of handling multi-modal data, scaling to multi-stage policies, and investigating deeper connections between large our of domain models and robot learning.
 
\bibliography{example_paper}
\bibliographystyle{icml2023} 
% APPENDIX 
\newpage
\appendix
\onecolumn

\section{Appendix}
% % \subsection{Additional Experiment Details}
% Videos are in an anonymized website \url{https://sites.google.com/view/cacti-anonymous/home}
 
\subsection{Implementation details on the framework's four stages}
\subsubsection{Stage 1}
\label{sec:stage1details}
In simulation, during the collect phase, we obtain 100 expert policies per task, corresponding to different layouts, hence a total of 1800 task and layout specific policies, which can be replayed in stage 2. For the real robot environment, we collect 8 trajectories per task through kinesthetic demonstration, so a total of 80 expert trajectories, which can be replayed. 

\subsubsection{Stage 2}
\label{sec:stage2details}

For augmenting the real-robot kinesthetic demos collected by experts, we replay the trajectories while varying different attributes of the scene, and recording per-timestep image observations during the replays.
The visual augmentations in the real-robot setting correspond to color jitters of the observation images. In addition, we incorporate three different semantic augmentations. The first is action noise during replays to ensure wider coverage in mitigating covariate shift issues. Second, we manually shuffle the positions of distractor objects across the scene, and swap some objects in and out of the scene. Finally, we develop a novel method for incorporating automatic semantic scene variations, \textit{without physically modifying objects} in the scene.  We use latest advances in generative modeling~\cite{dalle2,stablediffusion} that lets us perform controlled scene re-generations. This is at the dataset level, and doesn't require additional robot operation hours. We specifically consider the open-sourced Stable Diffusion trained model~\cite{stablediffusion}, and run inference through it. The model takes as input an image of the scene, and a region for modification, specified in pixel coordinates. Controlled generation lets us keep the rest of the scene unchanged, and introduce new plausible objects in the region specified. By automating this process, we can obtain several visually augmented demos with zero extra human effort for data collection. Fig. 7 shows a visualization of what controlled generation looks like for a scene from our real robot environment. The generated images place plausible objects like mugs, cups, and glasses on locations of the white-colored table that are unoccupied. 

\subsubsection{Stage 3}
\label{sec:stage3details}
The pre-trained visual representations for R3M are obtained through training on egocentric human videos~\cite{ego4d}, with a combination of time-contrastive loss, and losses for video-language alignment. We use the exact pre-trained model from the original paper, and do not introduce any additional loss for fine-tuning with our own collected data. Fine-tuning simply corresponds to backpropagating through the layers of the pre-trained encoder to update its weights, while performing imitation learning in stage 4.

\subsubsection{Stage 3}
\label{sec:stage4details}

For visual goals, the embeddings obtained from stage 3 are 1024x1 dimensional, and are concatnetaed with the observation embedding, which is also of the same dimensions, is concatenated, before feeding the concatenated vector to the policy MLP. In additon, we also concatenate the roobt joint velocity, and joint pose vectors (each of dimension 8x1), so the combined embedding that goes as input to the policy MLP is of dimension 2064x1. The output of the policy MLP is a mean and standard deviation vector, such that they represent a Gaussian action distribution of 8x1 dimension.

\subsection{Experiment Details on Real robot environment}
\label{app:real_exp}
\begin{figure}
    \centering
    \includegraphics[width=0.6\textwidth]{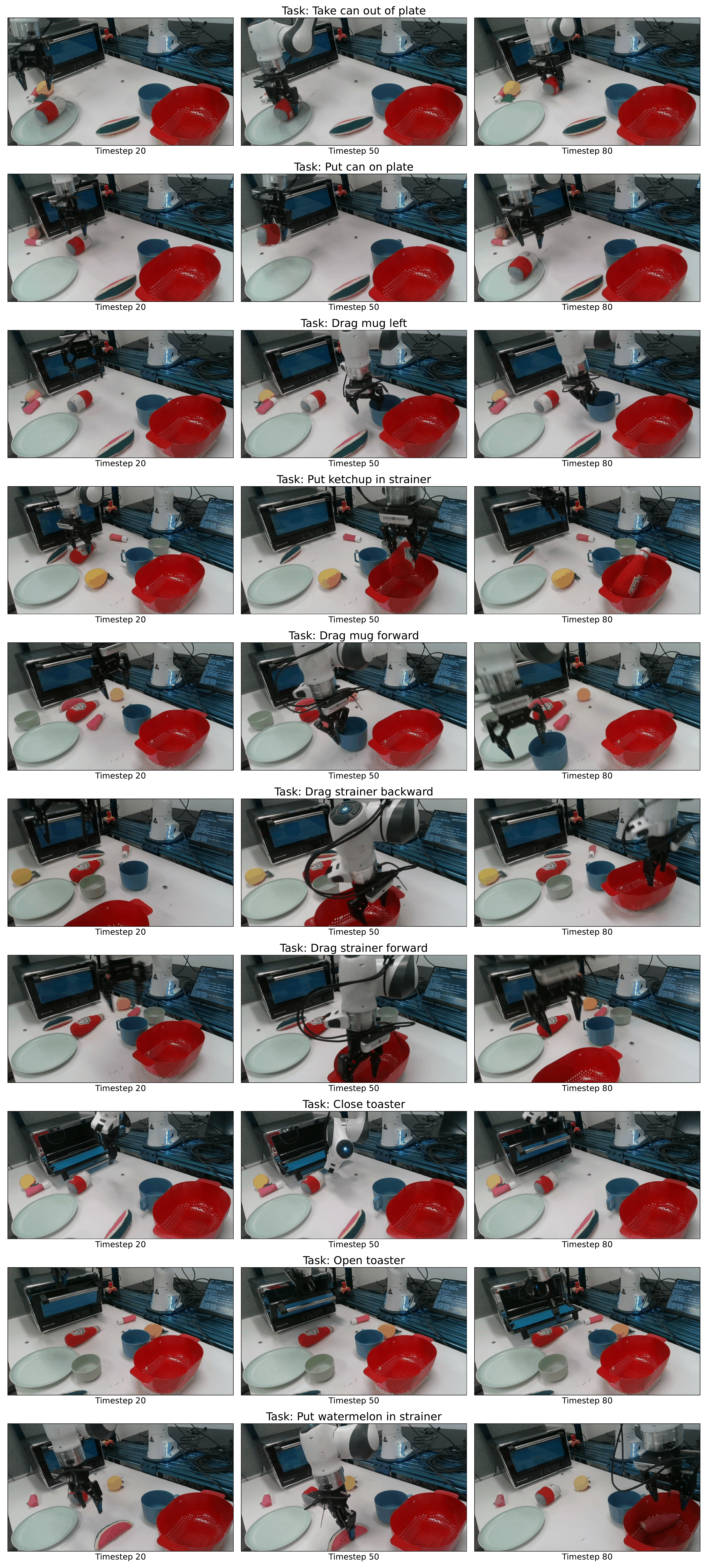}
    \caption{Visualizaition of all 10 training tasks in the training dataset from our real robot experiments. Each row of the above figure corresponds to one episode from one training task, where the three presented frames are sub-sampled from a 100-frames-long trajectory.}
    \label{fig:10task}
\end{figure}

\textbf{Task Setup}
The workspace contains a Franka Emika Panda robot arm, which is placed in front of a white tabletop with toy household objects, such as fruits and kitchen utensils. Based on these objects, we define each task to be the manipulation of an object from an initial location to a goal location: dragging mugs and strainers, picking and placing various objects, and opening and closing a toaster oven (please refer to Appendix Fig.~\ref{fig:10task} for visualizations of all 10 tasks). Each task episode in the training dataset contains 100 time-steps, each time-step contains RGB image observations of height 240 and width 424.

\textbf{Policy Deployment}
During deployment of a trained policy, each task is evaluated for a horizon length of 100 steps. At the beginning of each evaluation episode, a task name string is provided through user input to indicate which task to perform (e.g. ``drag\_mug"). This input text string gets embedded with a frozen text tokenizer (the same tokenizer that was used on \cite{r3m}) into a 768-dim task embedding. The same task embedding input is fixed across an episode, while the visual observations progress. We define an episode as success if the robot is able to move the target object to within a range of $3$ centimeter error from a task's desired goal location. To evaluate generalization, we randomly change the locations of distractor objects across the scene and induce organic variations in the initial positions of the target objects. We also introduce some random (previously unseen) distractor objects in the scene.

\subsection{Experiment Details on Simulation environment}
\label{app:sim_exp}
\textbf{Task Setup}
The environment contains 18 tasks, involving 8 main objects: four burner knobs, one light switch, one kettle, one cabinet with sliding door, one cabinet with a left and a right door, and one microwave. Some examples of the tasks are: turning on a light switch, opening the left door on cabinet, turning a knob. In parallel to the 18 semantic tasks, we also generate 100 random layouts that re-organizes the kitchen setup. For agent training, each combination of task and layout, which results in 1800 combinations in total, is represented by a 43-dimensional context embedding, and this embedding is used to condition a multi-task policy model. This task embedding is hand-crafted to contain the targeted object pose (unique to each task) and the object arrangement information (unique to each layout).
% goal configuration state of the task, and doesn't receive any proprioceptive state features of the objects in the per-timestep observation. 

\textbf{Policy Evaluation}
During evaluation, a task considered success if the final pose of the object at the end of an episode is within a certain error threshold of the goal pose of the object, and this error should stay within the threshould for more than 5 time-steps, which ensures the target object can reach a stable position and not get mistakenly counted as false positive success. To evaluate the policy's generalization to unseen scenarios, we additionally generate 10 more layouts in the simulation for each task, and randomly re-sample color, lighting, and texture properties for each of the 10 evaluation episodes. For our 100-layout policy, we evaluate 5 episodes for each of the 1800 task and layout combinations; for the 10-layout and 50-layout policies, we evaluate 10 episodes per training task and layout combination. 

% \subsection{Simulation Experiment Details}
% \subsubsection{Expert training}
% ffd
\end{document}